\documentclass[review]{elsarticle}
\pdfoutput=1

\usepackage{numcompress}
\usepackage{array}
\usepackage{hyperref}
\usepackage{amsmath}
\usepackage{mathrsfs}
\usepackage{graphics}
\usepackage{graphicx}
\usepackage{epstopdf}
\usepackage{indentfirst}
\usepackage{subfigure}
\usepackage{caption}
\usepackage{booktabs}
\usepackage{multirow}
\usepackage[english]{babel}
\usepackage{amsfonts,amssymb}
\usepackage[misc]{ifsym}
\usepackage{harpoon}
\usepackage{color}
\usepackage{geometry}
\usepackage{algorithm}
\usepackage{algorithmicx}
\usepackage{algpseudocode}
\renewcommand{\algorithmicrequire}{\textbf{Input:}}  
\renewcommand{\algorithmicensure}{\textbf{Output:}} 
\renewcommand{\algorithmicrepeat}{\textbf{Repeat:}} 
\renewcommand{\algorithmicuntil}{\textbf{Until:}} 

\journal{ISPRS Journal of Photogrammetry and Remote Sensing}

\begin{document}

\bibliographystyle{elsarticle-num}

\begin{frontmatter}

\title{An Effective Approach for Point Clouds Registration Based on the Hard and Soft Assignments}


\author{Congcong Jin\fnref{add1}}
\author[add1]{Jihua Zhu\corref{mycorrespondingauthor}}
\cortext[mycorrespondingauthor]{Corresponding author: Jihua Zhu}
\ead{zhujh@xjtu.edu.cn}
\author{Yaochen Li\fnref{add1}}
\author{Shaoyi Du\fnref{add1}}
\author{Zhongyu Li\fnref{add2}}
\author{Huimin Lu\fnref{add3}}
\address[add1]{School of Software Engineering, Xi'an Jiaotong University, China}
\address[add2]{University of North Carolina at Charlotte, USA}
\address[add3]{Kyushu Institute of Technology, Japan}




\begin{abstract}
For the registration of partially overlapping point clouds, this paper proposes an effective approach based on both the hard and soft assignments. Given two initially posed clouds, it firstly establishes the forward correspondence for each point in the data shape and calculates the value of binary variable, which can indicate whether this point correspondence is located in the overlapping areas or not. Then, it establishes the bilateral correspondence and computes bidirectional distances for each point in the overlapping areas. Based on the ratio of bidirectional distances, the exponential function is selected and utilized to calculate the probability value, which can indicate the reliability of the point correspondence. Subsequently, both the values of hard and soft assignments are embedded into the proposed objective function for registration of partially overlapping point clouds and a novel variant of ICP algorithm is proposed to obtain the optimal rigid transformation. The proposed approach can achieve good registration of point clouds, even when their overlap percentage is low. Experimental results tested on public data sets illustrate its superiority over previous approaches on accuracy and robustness.
\end{abstract}


\end{frontmatter}

\section{Introduction}
Aiming at building up correspondences and calculating the optimal transformation between two or more point clouds, point cloud registration is of great importance in 3D model reconstruction \cite{New15}, robot mapping \cite{Shiratori15,Yu15}, object recognition \cite{Abate07, Held16} and etc. As the most basic and essential perspective of multi-view registration, pair-wise registration has been devoted to by a large number of researchers.
Among the great deal of methods, the most original and popular one is the iterative closest point (ICP) algorithm proposed by P. J. Besl and N. D. McKay \cite{Besl92}, which iteratively builds up correspondences and calculates the rigid transformation by minimizing the Euclidean distance based residual error. However, this approach is time-consuming and only apply to two completely overlapping clouds.

In the past decade, a great deal of researchers have made efforts to improve the performance of the traditional ICP algorithm. To improve the low efficiency of ICP algorithm, Fitzgibbon \cite{Fitzgibbon03} employed the Levenberg---Marquardt algorithm to speed it up; Jost \cite{Jost03} proposed a coarse-to-fine multi-resolution technique with the neighbor search algorithm to improve its efficiency. For the partially overlapping problem neglected in original ICP, a straightforward solution is to exclude point correspondences with distances greater than the user-specified threshold \cite{Rusinkiewicz01}. After that, Godin \cite{Godin94} proposed weighting of point correspondences to assign lower weights to correspondences with greater point-to-point distances. But this approach is ineffective. Further, Chetverikov et al.\cite{Chet05} proposed the trimmed ICP (TrICP) algorithm, in which an overlapping parameter was introduced into the residual error. This approach can trim the outliers automatically and obtain accurate registration results for partially overlapping clouds, but it is time-consuming. To solve this problem, Phillips et al.\cite{Phillips07} proposed the fractional TrICP (FTrICP) algorithm to simultaneously compute the overlapping parameter and transformation in a much faster speed.

Although the aforementioned methods get substantial improvement in both robustness and efficiency, they still suffer either poor accuracy or local minimum problem. On one hand, to obtain more precise transformation, some probabilistic methods \cite{Granger02,Jian11,Myronenko10,Tsin04} were proposed for partial registration of point clouds. But the huge computational resources they require pose a great challenge to most application areas. On the other hand, to avoid the local minimum problem existing in many algorithms, Lee et al. \cite{Lee02} introduced invariant features into ICP to decrease the chance being trapped into local minimum. Besides, the genetic algorithm \cite{Lomonosov06, Zhu14} and particle filter \cite{Sandhu13} were also utilized to determine the optimal transformation.

In order to achieve more robust and accurate registration, some original methods based on the traditional ICP were proposed. For example, Zhu \cite{Zhu12} proposed the concept of bidirectional distances to solve the non-rigid registration of absolutely overlapping point clouds. Afterwards, backward distance measurement was introduced into affine ICP method by Ma et al.\cite{Ma13}. Although the bilateral correspondence based algorithms are more accurate than most other methods, their efficiency is unsatisfactory. Recently, correntropy \cite{Hasanbelliu14,Xu16} was introduced into ICP algorithm and can be solved by maximizing the correntropy between two point clouds. Although it can deal with rigid registration with noises and outliers, its robustness should be further improved for registration of cloud pair with low overlap percentage.

However, all the above methods can only be solved under an initial transformation that is close to the ground truth. Therefore, a group of researchers contribute to feature extraction to provide initial transformation for pair-wise registration. These algorithms are mainly classified into two categories, i.e., global feature based and local feature based algorithms. The objective of global feature based algorithms is to construct a set of features to encode the geometric properties of the entire 3D object, such as geometric 3D moments \cite{Paquet00}, shape distribution \cite{Osada02} and spherical harmonics \cite{Funkhouser03}. On the other hand, the objective of local feature based algorithms is to define a set of features to encode the characteristics of the local neighborhood of the feature points. For example, a set of point feature histograms was introduced to provide good starting points for ICP algorithm in \cite{Rusu08} and \cite{Rusu09}. Besides, Guo \cite{Guo13} designed a 3D object recognition algorithm based on his proposed local reference frame and rotational projection statistics. However, these algorithms usually suffer from either low descriptiveness or weak robustness.

Considering the aforementioned problems, we therefore propose an extended version of \cite{Zhu16} to achieve more robust and efficient registration. Compared with the original version, this approach introduces a variable of hard assignment into the objective function, in which each point in the data shape is assigned a binary value to indicate its probability as an inlier. Based on this step, the variable of soft assignment is further explored to illustrate the reliability of the established point correspondence. To compare the proposed approach with that without hard assignment, some experiments were performed on some real datasets and the experimental results demonstrate the advantages over the former method in efficiency, accuracy and robustness.

The rest of this paper is organized as the following: Section 2 briefly reviews the ICP algorithm. The motivation behind our proposed approach is explained in section 3.
Then in section 4, the proposed approach is illustrated in detail. After that is section 5, in which the superiority of the proposed method is further validated and some convincing experiments are demonstrated to compare the proposed algorithm with other related ones. At last, some conclusions are drawn in section 6.

\section{The ICP algorithm}

Given two point clouds, dnoted as the data shape $D =\{ {\vec d_i}\} _{i = 1}^{{N_d}}({N_d}\in{\mathbb{N}^3})$ and the model shape $M =\{ {\vec m_j}\} _{j = 1}^{{N_m}}({N_m}\in{\mathbb{N}^3})$, the goal of registration is to find the optimal rigid transformation $({\bf{R}},\vec t)$, with which $D$ can be in the best alignment with $M$. Accordingly, it can be formulated as the following least square (LS) problem:
\begin{equation}
\begin{array}{l}
\mathop {\min }\limits_{{\bf{R}},\vec t ,c\left( i \right) \in \left\{ {1, \ldots ,{N_m}} \right\}} \sum\limits_{i = 1}^{{N_d}} {\left\| {{\bf{R}}{{\vec d }_i} + \vec t  - {{\vec m }_{c\left( i \right)}}} \right\|} _2^2\\
\quad s.t.{{\bf{R}}^T}{\bf{R}} = {{\bf{I}}_{3 \times 3}},\det \left( {\bf{R}} \right) = 1
\end{array}
\label{eq:ICPObj}
\end{equation}
where ${\bf{R}} \in \mathbb{R}^{3\times 3}$ is the rotation matrix, $\vec t \in {\mathbb{R}^3}$ indicates the translation vector, ${\vec m_{c(i)}}$ represents the correspondence of ${\vec d_i}$ in the model shape, and ${\left\| \right\|}_2$ denotes $L_2$ norm. Eq. (\ref{eq:ICPObj}) can be solved by the ICP algorithm \cite{Besl92}, which achieves the rigid registration by iterations. Given an initial transformation $({\bf{R}_0},{\vec t_0})$, two steps are included in each iteration:

\begin{enumerate}[(1)]
\item According to $\left( {k - 1} \right)th$ transformation, establish correspondence for each point ${\vec d _i}$ in the data shape:
\begin{equation}
{c_k}\left( i \right) = \mathop {\arg \min }\limits_{j \in \left\{ {1,2, \ldots ,{N_m}} \right\}} {\left\| {{{\bf{R}}_{k - 1}}{{\vec d }_i} + {{\vec t }_{k - 1}} - {{\vec m }_j}} \right\|_2} \quad(i= 1,2, \ldots)
\label{eq:ICPCorr}
\end{equation}

\item Based on the established correspondences, calculated the transformation $\left( {{\bf{R}},\vec t } \right)$ by minimizing the following function:
\begin{equation}
\left( {{{\bf{R}}_k},{{\vec t }_k}} \right) = \mathop {\arg \min }\limits_{{\bf{R}},\vec t } \left( {\sum\limits_{i = 1}^{{N_d}} {\left\| {{\bf{R}}{{\vec d }_i} + \vec t  - {{\vec m }_{{c_k}\left( i \right)}}} \right\|_2^2} } \right)
\label{eq:ICPTrans}
\end{equation}

\end{enumerate}

Finally, the optimal transformation can be obtained by repeating steps (1)-(2) until some stop conditions are satisfied. Although the ICP algorithm has good performances for the rigid registration, it cannot achieve the registration of partially
overlapping point clouds.

\section{Motivation}

For point cloud registration, two issues should be carefully considered in practical applications: 1) The point cloud to be registered always contains non-overlapping areas, which should be confirmed and discarded during registration; 2) Even in the overlapping areas, there seldom exits real point correspondences due to the limited sensor resolution or noise.

\begin{figure*}[htbp]
\begin{minipage}[b]{.3\linewidth}
  \centering
  \centerline{\includegraphics[scale=0.5]{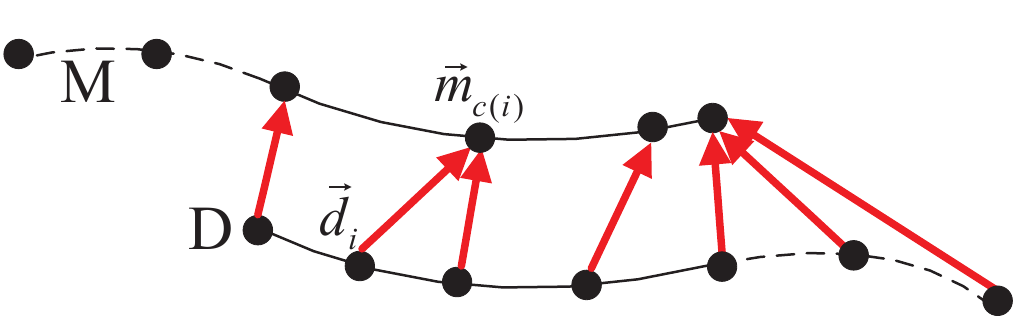}}
  \centerline{(a)}
\end{minipage}
\hfill
\begin{minipage}[b]{0.3\linewidth}
  \centering
  \centerline{\includegraphics[scale=0.5]{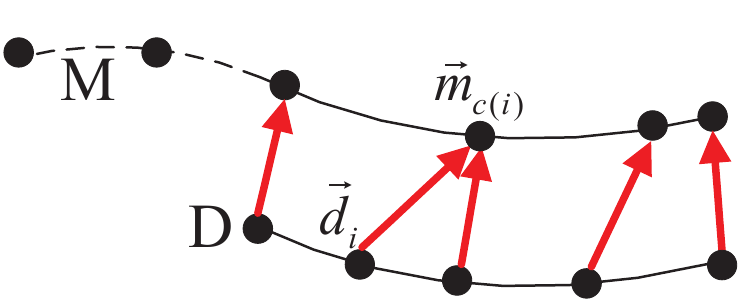}}
  \centerline{(b)}
\end{minipage}
\hfill
\begin{minipage}[b]{0.3\linewidth}
  \centering
  \centerline{\includegraphics[scale=0.5]{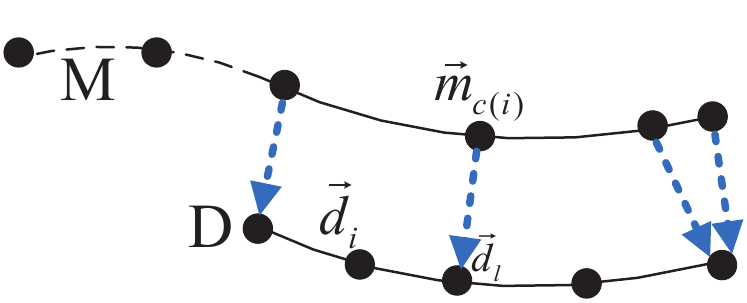}}
  \centerline{(c)}
\end{minipage}
\caption{ The framework of the proposed approach for registration of partially overlapping point clouds, where the solid curve indicates the overlapping areas, real and dashed line arrows denote the forward and backward distances, respectively. (a) Nearest neighbours with forward distances for each point in $D$. (b) According to the forward distance of each point, outliers can be discarded. (c) Correspondences with backward distances for some points in $M$, which are the nearest neighbours of points located in the overlapping areas of $D$.}\medskip
\label{fig:obse}
\end{figure*}

Given two partially overlapping point clouds, bilateral correspondences of one point in the data shape can be established by applying the search method of nearest neighbor twice. As shown in Fig. \ref{fig:obse}, for each point ${\vec d}_i $ in the data shape $D$, its nearest neighbour ${\vec m}_c(i) $ with the forward distance can be searched from the model shape $M$. Then, for each of these nearest neighbours in $M$, its correspondence ${\vec d}_l $ with the backward distance can also be searched from $D$. Consequently, two interesting phenomena can be reasonably discovered.

{\bfseries Observation 1: The forward distance of the point in overlapping areas is smaller than that of the point in non-overlapping areas:}
For one point of the data shape, if it is located in the overlapping areas, its correspondence can be found in the overlapping area of the model shape.
Therefore, the point pair of an established correspondence is very close to each other and the forward distance is always small. However, for one point located in the non-overlapping areas, it has no geometrically consistent correspondence. Hence, the point pair of an established correspondence is much far from each other and its forward distance is always larger than that of the point located in the overlapping area.
Accordingly, the value of forward distance can be viewed as the index to indicate whether one point is located in the overlapping areas or not.

{\bfseries Observation 2: The difference between forward and backward distances can denote the reliability of the established point correspondence.}
For the real point correspondence, the backward distance should be equal to its forward distance. In practice, the two points of one correspondence
do not have the same position on the object due to the limited senor resolution and noise. For one point in the data shape, the backward distance is usually shorter than the forward one. And there is only a small number of points whose forward distances are equal to their backward distances. As shown in Fig. \ref{fig:obse}(b) and (c), the less the difference between forward and backward distances is, the more likely this point pair of an established correspondence truly matches. In other words, with the increase of the difference, the reliability of point correspondence decreases. Therefore, the relationship of bidirectional distances can be viewed as an index to indicate whether the established point correspondence is reliable or not. If bidirectional distances of one point are almost equal, then its established correspondence can be viewed as the real correspondence with a higher probability. Otherwise the probability should accordingly decrease.

Actually, the points located in the non-overlapping areas are outliers, which should not be considered in registration. Besides, as the reliability of each point correspondence varies, different emphasis should be placed on individual point correspondence. Accordingly, a novel approach can be motivated to achieve the registration of partially overlapping point clouds.

\section{The proposed approach}
In this section, an effective approach is proposed for registration of partially overlapping point clouds.

\subsection{ICP algorithm based on the hard and soft assignments}

Based on the two observations, two variables can be introduced into point cloud registration: the variable for the hard assignment and the variable for the soft assignment. The former variable is a binary parameter utilized to indicate whether one point is an inlier or outlier, while the latter one is set to be a continuous variable between 0 and 1, which can denote the reliability of the established point correspondence. By introducing the former variable, outliers can be rejected during registration. Moreover, the latter variable is introduced to place more emphasis on these established point correspondences with high reliabilities.

Accordingly, a new objective function based on both hard and soft assignments can be proposed for registration of partially overlapping point clouds:
\begin{equation}
{\mathop {\min }\limits_{{\bf{R}},\vec t } \sum\limits_{i = 1}^{{N_d}} {\left\| {{\bf{R}}{{\vec d }_i} + \vec t  - {{\vec m }_{c\left( i \right)}}} \right\|} _2^2 \cdot {\omega _i} \cdot {p_i}}
\label{eq:ProObj}
\end{equation}
in which ${\omega _i}$ is the probability denoting ${\vec d _i}$ as an inlier and ${p_i}$ is the reliability of the established correspondence ${\vec d _i}$ and ${\vec m _{c\left( i \right)}}$. The binary probability ${\omega _i}$ is easy to determine since there is only two cases for one point, i.e. inlier or outlier. However, the definition of ${p_i}$ is much more complex considering its property as a real parameter. To take full advantage of observation 2, we turn to an exponential function based probability model to design accurate reliability of an established correspondence.

Noting that Eq. (\ref{eq:ProObj}) is an extension of Eq. (\ref{eq:ICPObj}), a similar solving approach to the traditional ICP algorithm can be adopted by iterating the following three steps:
\begin{enumerate}[(1)]
\item Based on $\left( {k - 1} \right)th$ transformation, establish bilateral correspondences for each point ${\vec d _i}$ in $D$:
\begin{equation}
{{c_k}\left( i \right) = \mathop {\arg \min }\limits_{j \in \left\{ {1,2, \ldots ,{N_m}} \right\}} {\left\| {{{\bf{R}}_{k - 1}}{{\vec d }_i} + {{\vec t }_{k - 1}} - {{\vec m }_j}} \right\|_2}}
\label{eq:ProBiCorrj}
\end{equation}
\begin{equation}
{l = \mathop {\arg \min }\limits_{j \in \{ 1,...,{N_d}\} } {\left\| {{{\vec m }_{c(i)}} - ({{\bf{R}}_{k - 1}}{{\vec d }_j} + {{\vec t }_{k - 1}})} \right\|_2}}
\label{eq:ProBiCorrl}
\end{equation}

\item Calculate the variables of hard assignment and soft assignment, respectively:
\begin{equation}
{{w_i} = \left\{ \begin{array}{l}
1,{\rm{\quad if \quad}}{\vec d _i} \in {D_\xi }\\
0,{\rm{\quad otherwise}}
\end{array} \right.}
\label{eq:hardass}
\end{equation}
\begin{equation}
{{p_i} = {e^{ - \gamma ({\rho _i} - 1)}}}
\label{eq:softass}
\end{equation}
in which ${D_\xi }$ is a subset of the data shape representing the overlapping areas in $D$, ${\rho _i}$ is the ratio of bidirectional distances and $\gamma $ is a preset parameter.

\item Update the rigid transformation by minimizing the objective function:
\begin{equation}
{({{\bf{R}}_k},{\vec t _k}) = \mathop {\arg \min }\limits_{{\bf{R}},\vec t } (\sum\limits_{i = 1}^{{N_d}} {\left\| {{\bf{R}}{{\vec d }_i} + \vec t  - {{\vec m }_{c\left( i \right)}}} \right\|} _2^2 \cdot {\omega _i} \cdot {p_i})}
\label{eq:ProTrans}
\end{equation}

\end{enumerate}

Under the given initial transformation, the optimal solution can be obtained by iteratively performing the above three steps until convergence criteria are satisfied.

\subsection{Establishing the bilateral correspondences}
In this paper, the $k-d$ tree based nearest neighbor searching method is adopted to find correspondences. For the transformed point $({{\bf{R}}_{k - 1}}{\vec d _i} + {\vec t _{k - 1}})$ in Eq. (\ref{eq:ProBiCorrj}), its nearest neighbor can be searched from the model shape $M$ in each iteration. Therefore, the $k-d$ tree for $M$ can be built before starting iteration and it only requires to be built once. However, the $k-d$ tree for the transformed data shape ${T_{k - 1}}(D)$ requires be continuously rebuilt in each iteration since the nearest neighbor for ${\vec m _{c(i)}}$ is searched from the transformed data shape each time, resulting in low efficiency of the whole approach. To improve efficiency, a modified version of Eq. (\ref{eq:ProBiCorrl}) is illustrated as:
\begin{equation}
{l = \mathop {\arg \min }\limits_{i \in \{ 1,...,{N_d}\} } {\left\| {{\bf{R}}_{k - 1}^{ - 1}{{\vec m }_{c(i)}} - {\bf{R}}_{k - 1}^{ - 1}{{\vec t }_{k - 1}} - {{\vec d }_i}} \right\|_2}}
\label{eq:ModiCorrl}
\end{equation}

Obviously, Eq. (\ref{eq:ProBiCorrl}) and (\ref{eq:ModiCorrl}) can achieve the same result. In this way, the search process can be accomplished through the following two steps: firstly, the $k-d$ tree based search method is adopted to find the nearest neighbor ${\vec m _{c(i)}}$ for each point in $D$; subsequently, for all the transformed nearest neighbors $({\bf{R}}_{k - 1}^{ - 1}{\vec m _{c(i)}} - {\bf{R}}_{k - 1}^{ - 1}{\vec t _{k - 1}})$, their corresponding nearest neighbors are searched from the data shape. Therefore, the $k-d$ tree for $D$ can be built only once before starting iteration.

In practical implementation, there is no need establishing bilateral correspondences for every point in $D$. Since the hard assignment has excluded most outliers from registration, we can only search bilateral correspondences for each inlier in the data shape, as shown in Fig. 1(b) and (c). Therefore, computational efficiency can be improved significantly.

\subsection{Calculating the variables of hard and soft assignments}
The aim of introducing hard assignment into registration is to assign a binary value for each point in $D$ to indicate its probability as an inlier. Considering that several parameters need determining in Eq. (\ref{eq:ProObj}), we thus turn to a simpler but similarly accurate method---the TrICP algorithm---for solution, in which the objective function is formulated as:
\begin{equation}
{\begin{array}{l}
\mathop {\min }\limits_{\xi ,{\bf{R}},\vec t } \left( {\frac{1}{{\left| {{D_\xi }} \right|{\xi ^{1 + \lambda }}}}\sum\limits_{{{\vec d }_i} \in {D_\xi }} {\left\| {{\bf{R}}{{\vec d }_i} + \vec t  - {{\vec m }_{c(i)}}} \right\|_2^2} } \right)\\
\quad s.t.{\rm{  }}{{\bf{R}}^T}{\bf{R}} = {{\bf{I}}_3},{\rm{  det}}\left( {\bf{R}} \right) = 1\\
\quad {\rm{     }}\xi  \in \left[ {{\xi _{\min }},1} \right],{\rm{ }}{{\rm{D}}_\xi } \subseteq D,{\rm{  }}\left| {{D_\xi }} \right| = \xi \left| D \right|
\end{array}}
\label{eq:TrICPObj}
\end{equation}
where $\lambda $ is a preset parameter, $\left|  \cdot  \right|$ denotes the cardinality of a set and $\xi $ is the overlapping percentage. Accordingly, those points that minimize the above function are viewed as inliers while others are outliers.

On the basis of \cite{Phillips07}, Eq. (\ref{eq:TrICPObj}) can be solved in a sequence processing manner by sorting the point correspondences $\left\{ {{{\vec m }_{c(i)}},({{\bf{R}}_{k - 1}}{{\vec d }_i} + {{\vec t }_{k - 1}})} \right\}_{c(i) = 1}^{{N_m}}$ according to their distances in ascending order and calculating
\begin{equation}
{\psi ({D_\xi }) = \frac{1}{{\left| {{D_\xi }} \right|{\xi ^{1 + \lambda }}}}\sum\limits_{{{\vec d }_i} \in {D_\xi }} {\left\| {{{\bf{R}}_{k - 1}}{{\vec d }_i} + {{\vec t }_{k - 1}} - {{\vec m }_{c(i)}}} \right\|_2^2} }
\label{eq:TrICPSol}
\end{equation}
At each iteration, a pair of sorted points is added to compute the corresponding value of $\psi ({D_\xi })$. By traversing all sorted point correspondences, it is easy to obtain the minimum value $\psi ({D_\xi })$ that corresponds to the optimal overlapping areas ${D_\xi }$. Subsequently, points belonging to ${D_\xi }$ are assigned the probability $1$.

Considering the property of the difference between forward and backward distances for a point in the data shape, we therefore adopt Eq. (\ref{eq:softass}) as the variable of soft assignment. We originally design the ${\rho _i}$ as the ratio of bidirectional distances, which is formulated as:
\begin{equation}
{{\rho _i} = \frac{{{{\left\| {{{\vec d }_i} - {{\vec m }_{c(i)}}} \right\|}_2}}}{{{{\left\| {{{\vec m }_{c(i)}} - {{\vec d }_l}} \right\|}_2}}}}
\label{eq:DefRho}
\end{equation}
in which ${\vec d _l}$ is the backward correspondence of ${\vec m _{c(i)}}$, ${\vec d _i} = ({{\bf{R}}_{k - 1}}{\vec d _i} + {\vec t _{k - 1}})$ and ${\vec d _l} = ({{\bf{R}}_{k - 1}}{\vec d _l} + {\vec t _{k - 1}})$.

Usually, the ratio of bidirectional distances is expected to meet the condition ${\rho _i} \in [1, + \infty )$. Sometimes, the point correspondence established by Eq. (\ref{eq:ProBiCorrj}) or Eq. (\ref{eq:ProBiCorrl}) may be coincident due to the round-off effect. In that case, ${\rho _i}$ will be illegal. Therefore, Eq. (\ref{eq:DefRho}) should be modified as follows:
\begin{equation}
{{\rho _i} = \frac{{{{\left\| {{{\vec d }_i} - {{\vec m }_{c(i)}}} \right\|}_2} + \delta }}{{{{\left\| {{{\vec m }_{c(i)}} - {{\vec d }_l}} \right\|}_2} + \delta }}}
\label{eq:ModiDefRho}
\end{equation}
where $\delta $ is a small positive parameter.

In practical applications, the effects of hard and soft assignments rely on parameters such as $\lambda $ and $\gamma$. For accurate registration, we want to select as many as inliers, so some outliers may be included into registration. To eliminate the disturbance, the parameter $\gamma$ can be adjusted to a smaller value to give hard penalty to these selected outliers.

\subsection{Computation of the rigid transformation}
To compute the rigid transformation $({\bf{R}},\vec t)$, the following derivation presents the closed-form solution for this question.

Substitute ${q_i}$ for the product of two variables of hard and soft assignments:
\begin{equation}
{{q_i} = {\omega _i} \cdot {p_i}}
\label{eq:SubstiMul}
\end{equation}
Extend the quadratic-formed objective function depicted in Eq. (\ref{eq:ProObj}) as following:

\begin{equation}
\begin{array}{l}
 J({\bf{R}},\vec t) = \sum\limits_{i = 1}^{{N_d}} {[{{({\bf{R}}{{\vec d}_i} - {{\vec m }_{c(i)}})}^T}({\bf{R}}{{\vec d}_i} - {{\vec m }_{c(i)}}){q_i}}  \\
  + 2{{\vec t}^T}({\bf{R}}{{\vec d}_i} - {{\vec m }_{c(i)}}) {q_i} +  {{\vec t}^{\rm{T}}} \cdot {\vec t} \cdot {q_i}] \\
 \end{array}
 \label{eq:J}
\end{equation}
Take the derivative of $J$ with respective to $\vec t$, the following result can be acquired:
\begin{equation}
\frac{{\partial J}}{{\partial {\vec t}}} = \sum\limits_{i = 1}^{N_d} {\left[ {2({\bf{R}}{{\vec d}_i} - {{\vec m }_{c(i)}}){q_i} + 2\vec t \cdot {q_i}} \right]}
\label{eq:Jt}
\end{equation}
Let $\frac{{\partial J}}{{\partial {\vec t}}}=0$, the translate vector can be obtained as:
\begin{equation}
{\vec t}_k =  - \frac{{\sum\limits_{i = 1}^{N_d} {({\bf{R}}{{\vec d}_i} - {{\vec m }_{c(i)}}){q_i}} }}{{\sum\limits_{i = 0}^{N_d} {{q_i}} }}
\label{eq:OptT}
\end{equation}

In Eq. (\ref{eq:J}), $\vec t$ can be replaced by Eq. (\ref{eq:OptT}) and the objective function can be simplified as:
\begin{equation}
J({\bf{R}}) = \sum\limits_{i = 1}^{{N_d}} {\left\| {{\bf{R}}{{\vec x}_i} - {{\vec y}_i}} \right\|_2^2}  \cdot {q_i}
\end{equation}
where ${\vec x_i} = {\vec d_i} - {{\sum\limits_{i = 1}^{{N_d}} {{q_i}} {{\vec d}_i}} \mathord{\left/
 {\vphantom {{\sum\limits_{i = 1}^{{N_d}} {{q_i}} {{\vec d}_i}} {\sum\limits_{i = 1}^{{N_d}} {{q_i}} }}} \right.
 \kern-\nulldelimiterspace} {\sum\limits_{i = 1}^{{N_d}} {{q_i}} }}$ and ${\vec y_i} = {\vec m_{c(i)}} - {{\sum\limits_{i = 1}^{{N_d}} {{q_i}} {{\vec m}_{c(i)}}} \mathord{\left/
 {\vphantom {{\sum\limits_{i = 1}^{{N_d}} {{q_i}} {{\vec m}_{c(i)}}} {\sum\limits_{i = 1}^{{N_d}} {{q_i}} }}} \right.
 \kern-\nulldelimiterspace} {\sum\limits_{i = 1}^{{N_d}} {{q_i}} }}$. Subsequently, the rotation matrix can be calculated by minimizing the function $J(\bf{R})$, which can be expanded as:
 \begin{equation}
\begin{array}{l}
{\bf{R}}= \mathop {\arg \min }\limits_{\bf{R}}  {\sum\limits_{i = 1}^{{N_d}} ({{\vec x}_i^T{{\vec x}_i} + {\vec y}_i^T{{\vec y}_i} - 2{\vec x}_i^T{\bf{R}}{{\vec y}_i}} })\cdot {q_i} \\
\quad  = \mathop {{\mathop{\arg \max}\nolimits} }\limits_{\bf{R}} \sum\limits_{i = 1}^{{N_d}} ({{\vec x}_ i^T{\bf{R}}{{\vec y}_i}}{q_i})
\end{array}
\label{eq:Arun}
 \end{equation}

Taking advantage of the optimization for Eq. (\ref{eq:Arun}) solved by Myronenko \cite{Myronenko09}, the conclusion can be directly presented as:

1) Compute the $3 \times 3$ matrix $\bf{H}$ and its singular value decomposition (SVD) results:
  \begin{equation}
  {\bf{H}} = \sum\limits_{i = 1}^{{N_d}} {{{\vec y}_i}} {q_i}\vec x_i^T\end{equation}
  \begin{equation}
[{\bf{U}},\Lambda ,{\bf{V}}] = {\mathop{\rm SVD}\nolimits} ({\bf{H}})
\end{equation}

2) Calculate the rotation matrix:
\begin{equation}
{\bf{R}} = {\bf{V}}{\bf{D}}{{\bf{U}}^T}
\end{equation}
where ${\bf{D}} = diag(1,1,\det ({\bf{V}}{{\bf{U}}^T}))$.

\subsection{Algorithm implementation}

According to the above analysis and derivation, the implementation of the proposed approach can be summarized in Algorithm 1.
\floatname{algorithm}{Algorithm}
\renewcommand{\algorithmicrequire}{\textbf{Input:}}
\renewcommand{\algorithmicensure}{\textbf{Output:}}
\renewcommand{\algorithmicrepeat}{\textbf{Repeat:}}
\renewcommand{\algorithmicuntil}{\textbf{Until:}}

    \begin{algorithm}[htbp]
        \caption{: ICP algorithm based on the hard and soft assignments}
        \begin{algorithmic}[1] 
            \Require Initial transformation $({{\bf{R}}_0},{\vec t _0})$
            \Ensure Refined transformation $({{\bf{R}}},{\vec t})$\\
            Create $k-d$ trees for $M$ and $D$, respectively;
            \Repeat
              \State Establish bilateral correspondences for each point ${\vec d _i}$ in $D$ according to Eq. (\ref{eq:ProBiCorrj}) and (\ref{eq:ProBiCorrl});
              \State Calculate the value of ${\omega _i}$ and ${p_i}$ based on Eq. (\ref{eq:hardass}) and (\ref{eq:softass}), respectively;
              \State Compute the rigid transformation according to subsection $4.1.$;
            \Until Stop criteria are satisfied.
        \end{algorithmic}
    \end{algorithm}

Given the initial parameters $({{\bf{R}}_0},{\vec t _0})$, the proposed approach can achieve registration of partially overlapping point clouds with good performances.

\section{Experimental results}
In this section, we firstly validated the necessity of the hard and soft assignments combined method for registration. Through a series of experimental results, the superiority of introducing hard assignment into registration is self-evident. Then in the second part, the proposed approach was compared with several methods both for data simulation and for experiment comparison. Finally, the proposed method was applied to some large-scale environmental point clouds. All of the following experiments adopt the $k-d$ tree based nearest neighbor searching method on one same computer.

\subsection{Validation}

\begin{figure*}[htbp]
  \centering
  \centerline{\includegraphics[scale=0.7]{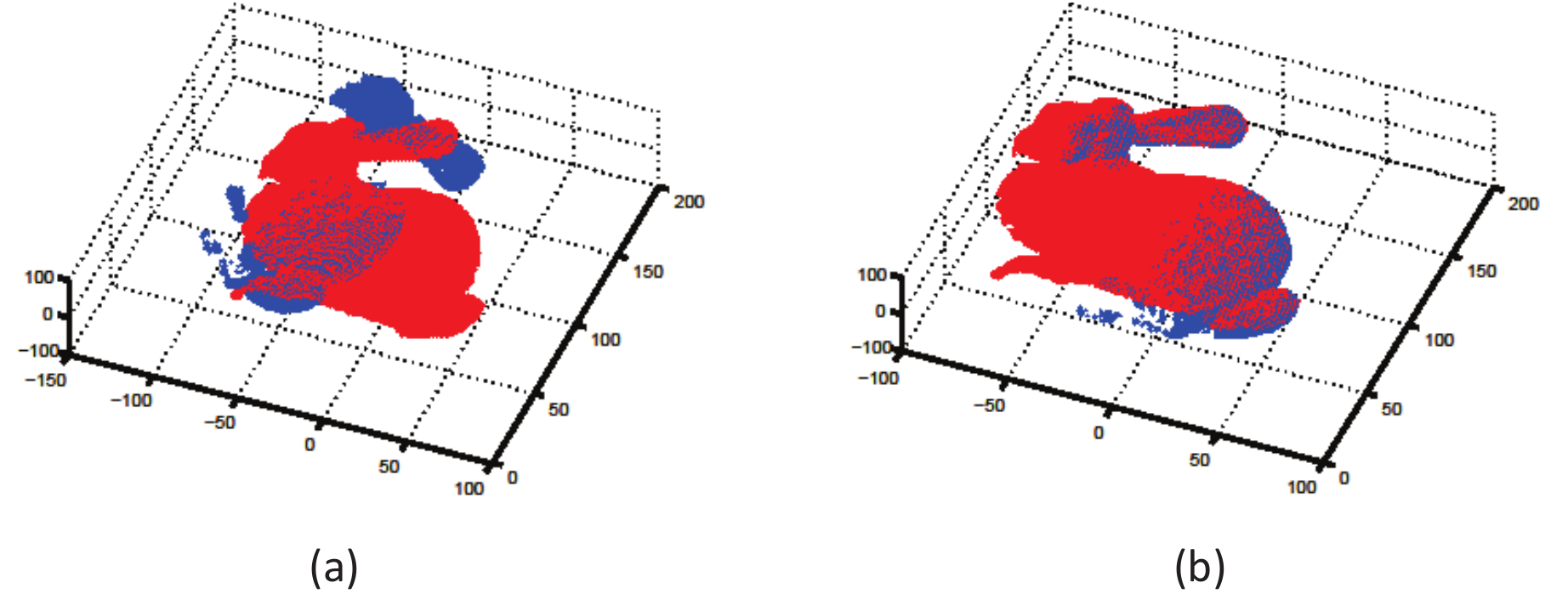}}

\caption{Registration comparison for a pair of point clouds. (a) Result using BdICP. (b) Result using the proposed approach.}\medskip

\label{fig:bunfail}
\end{figure*}

A significant advantage of introducing soft assignment into registration is the high accuracy it brings about. However, besides the low efficiency due to bilateral correspondences for all points in the data shape, it can not avoid convergence of local minimum neither. By introducing hard assignment to exclude outliers, this phenomenon can be alleviated remarkably. Fig. \ref{fig:bunfail} displays the registration results without and with the hard assignment strategy. For a pair of point clouds from the Stanford Bunny \cite{StanfordRep}, the algorithm using soft assignment converges to local minimum and results in failing registration, while our proposed method can obtain precise results, as shown in Fig. \ref{fig:bunfail} (a) and (b), respectively.

\begin{figure*}[t]
\centering

\centerline{\includegraphics[scale=0.5]{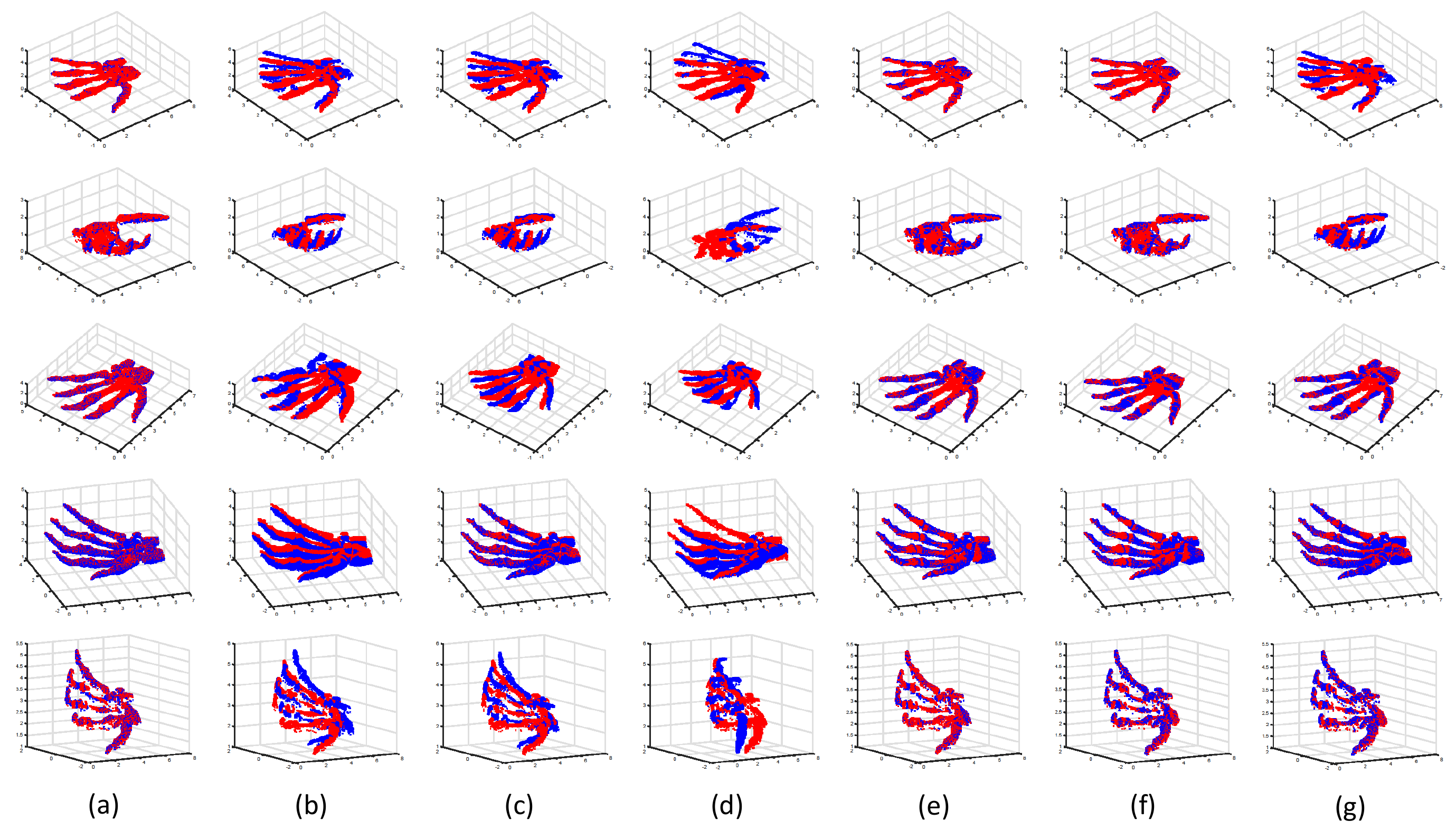}}

\caption{ Registration results for five pairs of clouds. (a) Registration results under ground truth. (b) Registration results of BdICP under Turb. 1. (c) Registration results of BdICP under Turb. 2. (d)Registration results of BdICP under Turb. 3. (e) Registration results of the proposed approach under Turb. 1. (f) Registration results of the proposed approach under Turb. 2. (g) Registration results of the proposed approach under Turb. 3.}\medskip
\label{fig:hand}

\end{figure*}

To further demonstrate the necessity of introducing hard assignment, the following experiments were designed to compare the performances without and with the hard assignment. In this experiment, five pairs of point clouds scanned from different viewpoints of one same object were randomly chosen. For each scan pair, the ground truth of transformation $\left( {{{\bf{R}}_g},{{\vec t }_g}} \right)$ was imposed by different perturbed transformations $\left( {{{\bf{R}}_r},{{\vec t }_r}} \right)$,  which were randomly drawn from the uniformation distribution $U\left[ { - 5\deg ,5\deg } \right]$, $U\left[ { - 5{\mathop{\rm d}\nolimits} ,5{\mathop{\rm d}\nolimits} } \right]$; $U\left[ { - 10\deg ,10\deg } \right]$, $U\left[ { - 5{\mathop{\rm d}\nolimits} ,5{\mathop{\rm d}\nolimits} } \right]$; and $U\left[ { - 10\deg ,10\deg } \right]$, $U\left[ { - 7{\mathop{\rm d}\nolimits} ,7{\mathop{\rm d}\nolimits} } \right]$ respectively, in which $d$ is the average point resolution of model shape. We respectively denote these turbulence as Turb. 1, Turb. 2 and Turb. 3. Taking the rigid transformation $\left( {{{\bf{R}}_r} \cdot {{\bf{R}}_g},{{\vec t }_r} + {{\vec t }_g}} \right)$ as initial transformation, the estimated transformation $\left( {{{\bf{R}}_m},{{\vec t }_m}} \right)$ can be obtained accordingly. For comparing accuracy of both approaches, relative errors are defined as: ${\varepsilon _{\bf{R}}} = {\left\| {{{\bf{R}}_m} - {{\bf{R}}_g}} \right\|_F}$ and ${\varepsilon _{\vec t }} = \frac{{{{\left\| {{{\vec t }_m} - {{\vec t }_g}} \right\|}_2}}}{d}$. Table 1 records the overlapping percentage of each scan pair and the runtime for the methods before and after introducing hard assignment under different turbulences. To demonstrate the results figuratively, Fig. \ref{fig:hand} displays the registration results.

\begin{table}[htbp]
\centering
\caption{Efficiency comparison of BdICP and the proposed approach for different overlapping point clouds.}
\label{my-label}
\scalebox{1}{
\begin{tabular}{c|c|r|r|r|r|r|r}
\hline
\multirow{2}*{No. of scan pair}&     \multirow{2}*{Overlap}&  \multicolumn{2}{c|}{Turb. 1}&  \multicolumn{2}{c|}{Turb. 2}& \multicolumn{2}{c}{Turb. 3} \\ \cline{3-8}
                      &                                             &                                             BdICP&    Ours&   BdICP&    Ours&   BdICP&   Ours  \\ \hline
                     1&                                                                       47.11\%&            1.9889&   0.4696&  1.6356&  0.5643&  6.4039&  2.6563 \\ \hline
                     2&                                                                       54.92\%&            2.0694&   0.6242&  1.4216&  0.8189&  6.7972&  2.9957 \\ \hline
                     3&                                                                       63.19\%&            2.3623&   1.5701&  2.7494&  1.1624&  4.4361&  2.1182 \\ \hline
                     4&                                                                       72.63\%&            4.0802&   4.1701&  5.7047&  3.0769&  4.1523&  2.0203 \\ \hline
                     5&                                                                       81.05\%&            1.7954&   0.8091&  1.3961&  0.5694&  2.4287&  2.2407 \\ \hline
\end{tabular} }
\end{table}
As demonstrated in Fig. \ref{fig:hand}, in each trial, the relative errors generated from the proposed method are smaller than that without using hard assignment. Although the errors of both methods will increase with the raise of turbulence, the proposed method merely gains a moderate increase while the method solely using soft assignment has an apparent augmentation. As for the soft assignment approach, the registration results for higher overlapping cloud pairs were not always better than that for lower overlapping cloud pairs under the same turbulence. This is because the approach without using hard assignment is easy to be trapped into local minimum and thus always results in failing registration. Best illustrated in low overlapping situations, the proposed algorithm could effectively avoid the local minimum problem. Besides, Table 1 demonstrates a distinct comparison about efficiency. As it shows, the efficiency of the proposed method showed an obvious superiority over merely soft assignment method, especially when the overlap percentage was low. With the increase of turbulences, efficiency of both methods will decrease, but the proposed method still outperformed the method without using hard assignment. This high efficiency owes to the hard assignment strategy, which excludes most outliers from establishing bilateral correspondences and computing the ratio of bilateral distances. Obviously, the proposed method using both hard and soft assignments is superior to that merely using soft assignment in efficiency, accuracy and robustness.

\subsection{Comparison}
To testify its superior performances, the proposed approach was compared with the fractional TrICP algorithm (FTrICP), ICP algorithm with weighting of pairs (wICP) \cite{Rusinkiewicz01} and the correntropy ICP algorithm (CtICP) on several real datasets. For providing solid proof about their performances, the experiments were conducted both for data simulation and for experiment comparison.

\subsubsection{Data simulation}

To perform the simulation comparison, a point cloud with $8171$ 3D points was selected from the Stanford Bunny. For generating the data shape, the following processes were implemented: The original shape was randomly deleted $5\% $ points and then cut down by one part, which included $n$ range points; then the remaining shape was added with Gaussian zero mean noise. For generating the model shape, the following two steps were implemented: Deleting $5\% $ points in the original shape and applying a randomly generated transformation $\left( {{{\bf{R}}_r},{{\vec t }_r}} \right)$ to the remaining shape; then cutting down the transformed shape by other $n$ range points. Accordingly, the overlap percentage of the two generated shapes is calculated as:
\begin{equation}
\xi  = {{0.95 \times \left( {0.95N - 2n} \right)} \mathord{\left/
 {\vphantom {{0.95 \times \left( {0.95N - 2n} \right)} {\left( {0.95N - n} \right)}}} \right.
 \kern-\nulldelimiterspace} {\left( {0.95N - n} \right)}}
\end{equation}
in which $N$ denotes the number of points of the original shape. To simulate the registration results under different overlap percentages, the value $n$ can be accordingly changed. Thus we can obtain a set of registration results for all competed approaches.

\begin{figure*}[!t]
\begin{minipage}[b]{.5\linewidth}
  \centering
  \centerline{\includegraphics[scale=0.4]{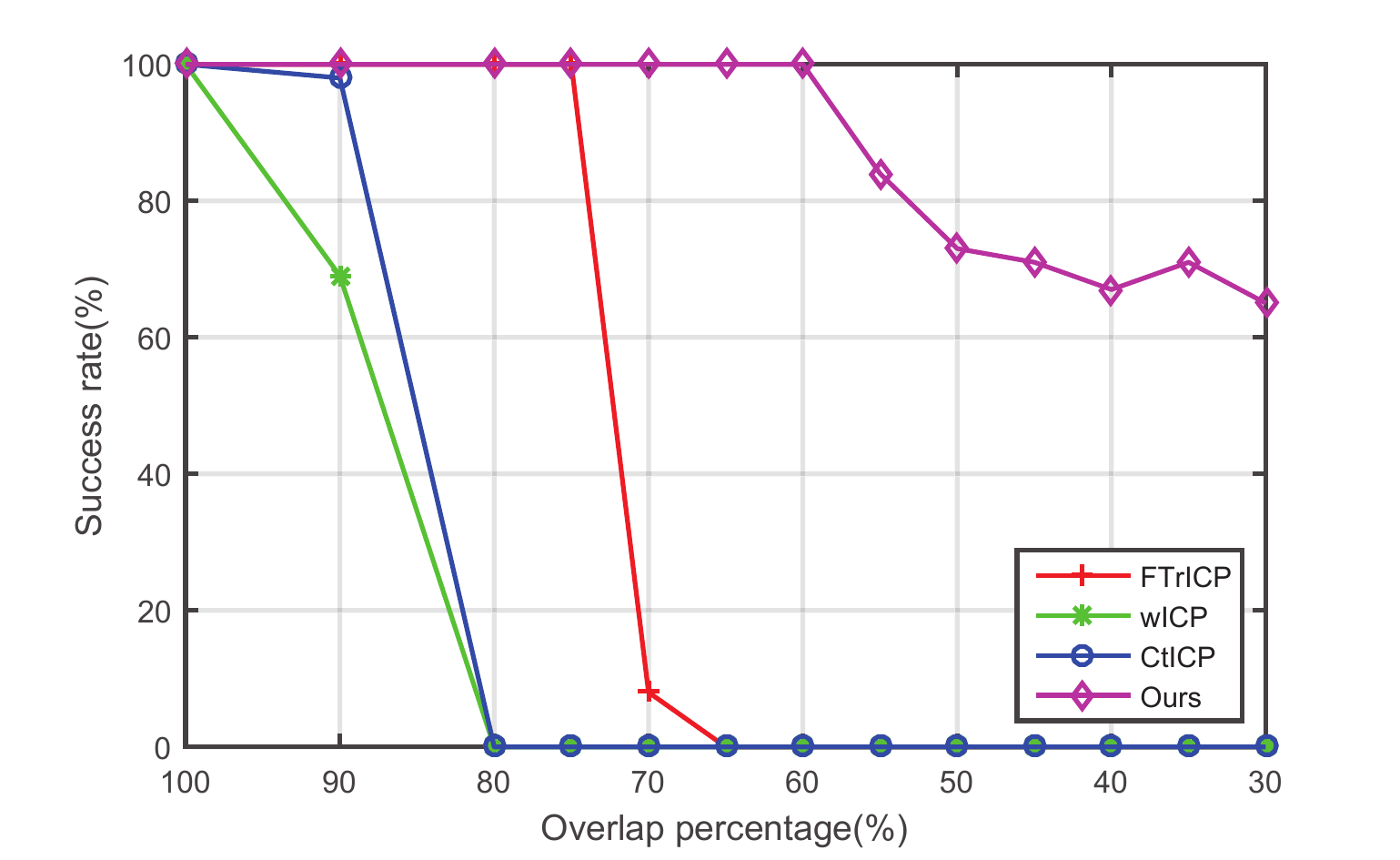}}
  \centerline{(a)}
\end{minipage}
\hfill
\begin{minipage}[b]{0.5\linewidth}
  \centering
  \centerline{\includegraphics[scale=0.4]{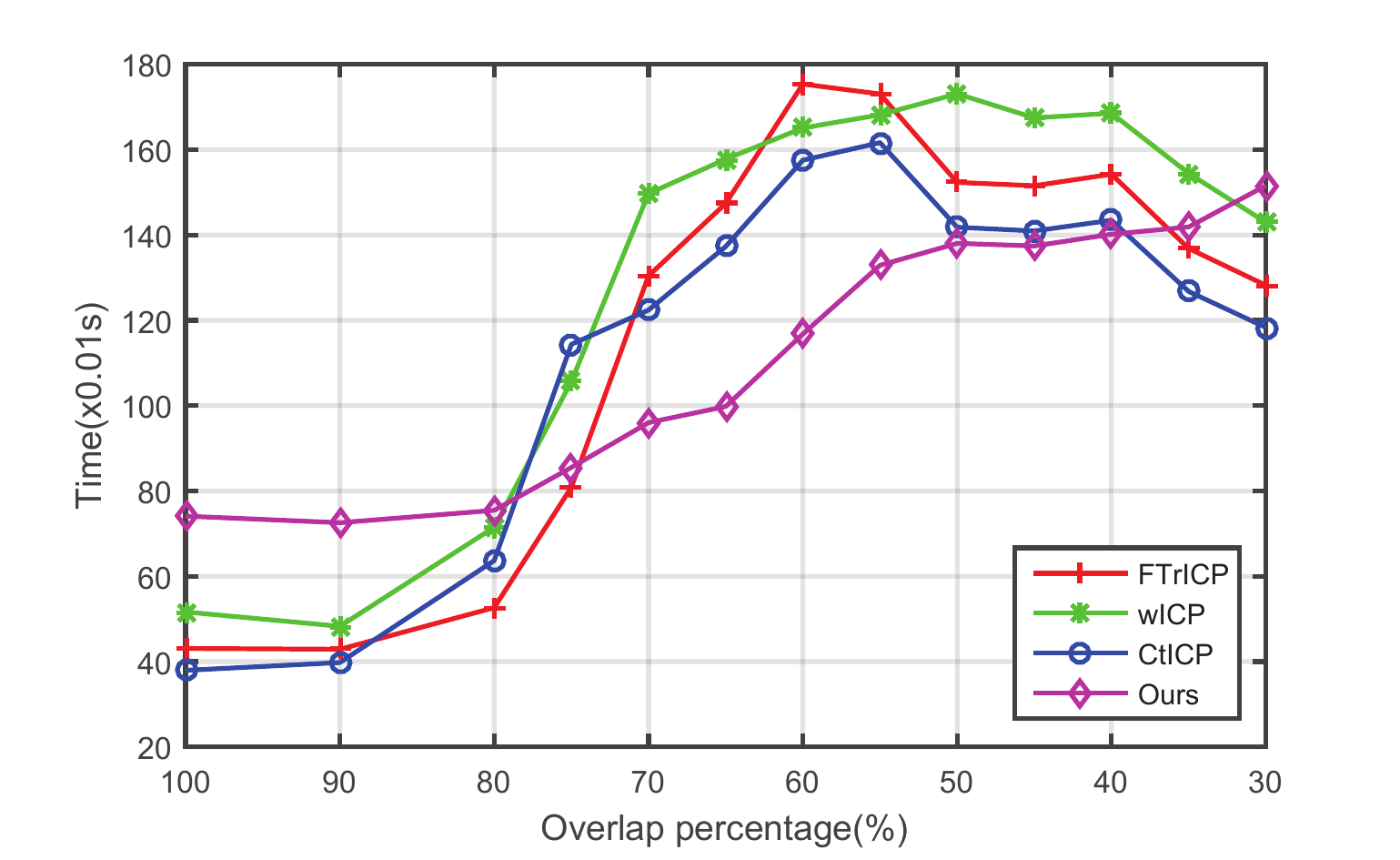}}
  \centerline{(b)}
\end{minipage}

\caption{ Performance comparison under varied overlap percentages. (a) Success rate. (b) Average runtime.}\medskip
\label{fig:sim}
\end{figure*}

For comparison, relative errors are defined as ${\varepsilon _{\bf{R}}} = {\left\| {{{\bf{R}}_m} - {{\bf{R}}_r}} \right\|_F}$ and ${\varepsilon _{\vec t }} = {\left\| {{{\vec t }_m} - {{\vec t }_r}} \right\|_2}$, where $\left( {{{\bf{R}}_m},{{\vec t }_m}} \right)$  represents the estimated transformation, and ${\left\| {} \right\|_F}$ denotes Frobenius norm. Denote $d$ as the average point resolution of model shape, the registration process can be viewed as successful if and only if (${\varepsilon _{\bf{R}}} \le 0.01$ and ${\varepsilon _{\vec t }} \le d$). All approaches were tested on each cloud pairs with varied overlap percentages for $100$ Monte Carlo (MC) trials. To view it more intuitively, Fig. \ref{fig:sim} displays registration performances for all competed approaches.

As shown in Fig. \ref{fig:sim}(a), both wICP and CtICP are satisfactory for pretty highly overlapping clouds. This is because the correspondence distance threshold specified by user is mainly determined empirically and many outliers may be included into registration. The FTrICP algorithm can achieve more superior performance compared with the former two methods. But its robustness decreases quickly with the decrease of overlap percentage. This is a common phenomenon since many methods neglect the reliability of established correspondences. On the contrary, the proposed approach can still obtain precise registration results even though the overlap percentage is pretty small. Moreover, its robustness decreases quite slowly with the increase of outliers. Benefited from the hard assignment, this method can gain a slight raise on registration precision when the overlap percentage is small enough. Therefore, the proposed approach achieves the most robust registration among all the competed approaches. In terms of efficiency, Fig. \ref{fig:sim}(b) gives the best illustration. When the overlap percentage is pretty high, the proposed approach has the lowest efficiency since it needs to compute bilateral distances for most points. Once the overlap percentage is lower than $80\% $, its superiority over other approaches becomes distinct and even has the highest efficiency among all the competed approaches. This benefits from the hard assignment strategy, which has excluded most outliers from registration.

\subsubsection{Experiment comparison}

\begin{figure*}[!htb]
\centering
\centerline{\includegraphics[scale=0.6]{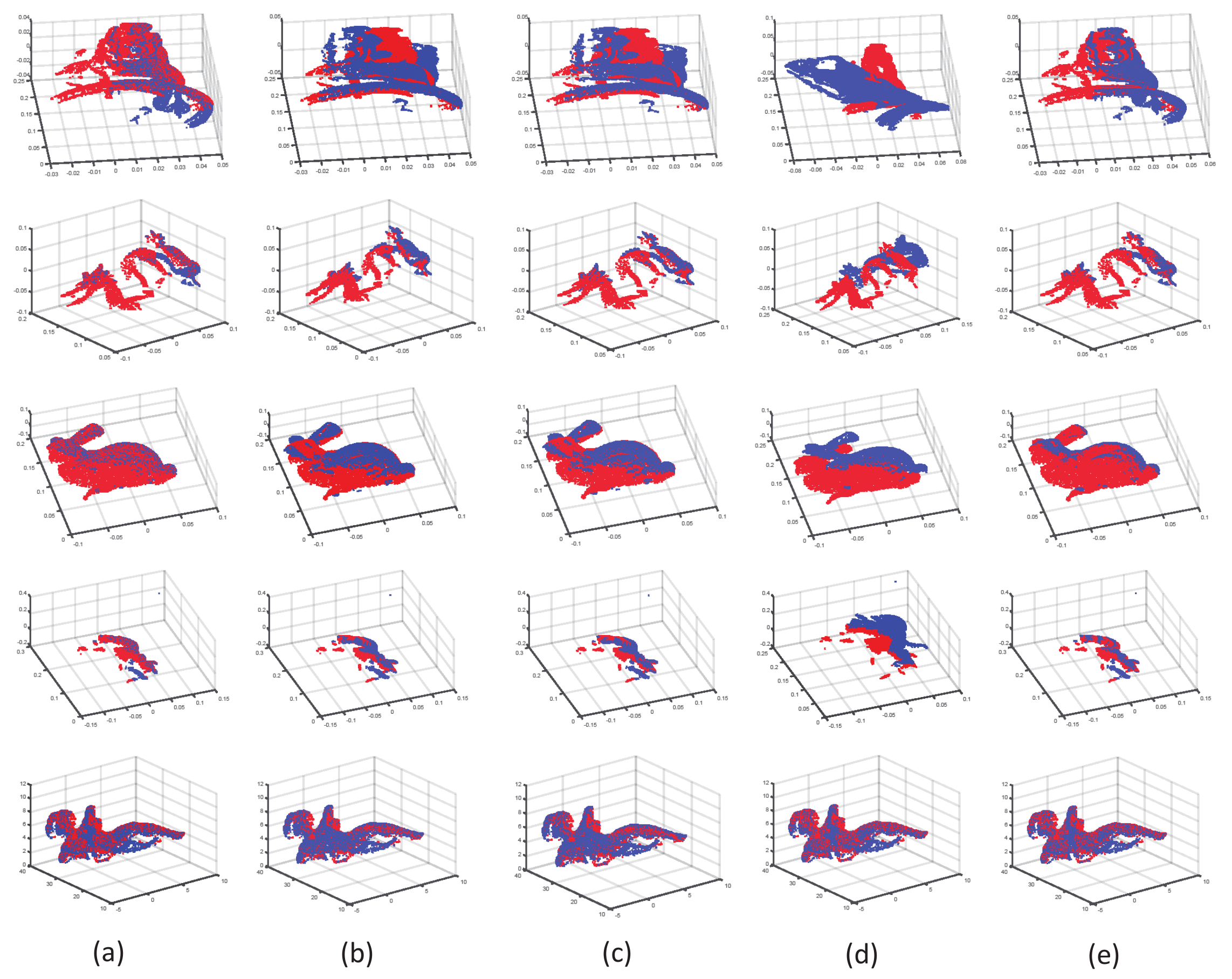}}

\caption{ Registration results via all competed approaches. (a) Results under ground truth. (b) FTrICP. (c)wICP. (d) CtICP. (e) Ours.}\medskip
\label{fig:setscom}

\end{figure*}

In this section, the proposed approach was compared with the aforementioned methods on several real datasets: Happy Buddha, Dragon, Bunny, Armadillo and Angel. Detailed information of each cloud pair was depicted in Table 2. In addition to point clouds information, the ground truth of rigid transformation $\left( {{{\bf{R}}_g},{{\vec t }_g}} \right)$ for each pair of cloud is also known.

Before registration, a perturbed rigid transformation $\left( {{{\bf{R}}_r},{{\vec t }_r}} \right)$ was imposed to the ground truth, in which the Euler angle and translate component were randomly drawn from the uniform distribution $U\left[ { - 5\deg ,5\deg } \right]$ and $U\left[ { - {\mathop{\rm d}\nolimits} ,{\mathop{\rm d}\nolimits} } \right]$ respectively. Here, the symbol $d$ denotes the average point resolution of the model shape. Taking the rigid transformation $\left( {{{\bf{R}}_r} \cdot {{\bf{R}}_g},{{\vec t }_r} + {{\vec t }_g}} \right)$ as initial parameter for all competed approaches, the estimated transformation $\left( {{{\bf{R}}_m},{{\vec t }_m}} \right)$ can be obtained accordingly. For comparing accuracy of each approach, relative errors are defined as: ${\varepsilon _{\bf{R}}} = {\left\| {{{\bf{R}}_m} - {{\bf{R}}_g}} \right\|_F}$ and ${\varepsilon _{\vec t }} = \frac{{{{\left\| {{{\vec t }_m} - {{\vec t }_g}} \right\|}_2}}}{d}$. All competed approaches were tested on each dataset with $20$ MC trials to eliminate random interference. To evaluate the performance of each approach, Table 2 demonstrates the relative error of rotation matrix, translation component and average runtime, respectively. To view this in a more intuitive way, Fig. \ref{fig:setscom} displays the registration results for all competed approaches.

As shown in Table 2 and Fig. \ref{fig:setscom}, all approaches can achieve precise registration on Angel, which has high overlap percentage.  Since it neglects outliers and relies highly on initial transformation, the CtICP algorithm obtained unsatisfactory results on the other four datasets. While FTrICP and wICP achieved more precise and efficient registration than CtICP on most datasets, they still fall behind our proposed approach by a large margin. As for the wICP algorithm, it neglects the nonoverlapping areas problem and searches a correspondence for each point in the data shape, which includes many outliers into registration. Although the FTrICP algorithm takes outliers into consideration, it deems that all the established correspondences are correct, thus may fail when registration are interfered by many outliers. The relative errors of ${\bf{R}}$ and ${\vec t }$ increase dramatically for the FTrICP, wICP and CtICP with the decrease of overlap percentage, but ours has a moderate rise profited from the combination of hard and soft assignments. Besides, the high efficiency of the proposed method resulted from the hard assignment also put it in an advantageous status. Based on these analysis, it is fair to draw that the proposed method is the most robust and efficient one among all these competed approaches.

\begin{table}[t]
\centering
\caption{Performance comparison of all competed approaches tested on five datasets.}
\label{my-label}
\scalebox{1}{
\begin{tabular}{c|c|c|c|c|c|c|c|c|c}
\hline
              Dataset&               Overlap&         {${N_m}$}&                   $d$&            {${N_d}$}&       Terms&                    FTrICP&          wICP&     CtICP&     Ours \\ \hline
\multirow{3}*{Buddha}&  \multirow{3}*{33.29\%}&  \multirow{3}*{12147}&      \multirow{3}*{0.0463}&     \multirow{3}*{15233}&    ${\varepsilon _{\bf{R}}}$&       1.0087&   	   1.2292&	    0.4599&	    0.2015 \\
                       &                        &                  &                      &                     &   ${\varepsilon _{\vec t }}$&       0.2560&          0.3470&      1.0240&     0.3394 \\
                       &                        &                  &                      &                     &                       $T(s)$&       5.5287&          8.4895&      4.8590&     0.3312 \\ \hline
  \multirow{3}*{Dragon}&  \multirow{3}*{48.36\%}&  \multirow{3}*{4418}& \multirow{3}*{0.0627}&          \multirow{3}*{2976}&    ${\varepsilon _{\bf{R}}}$&       0.2380&          0.1864&      0.7242&     0.1042 \\
                       &                        &                  &                      &                     &   ${\varepsilon _{\vec t }}$&       0.2752&          0.2362&      1.2914&     0.2423 \\
                       &                        &                  &                      &                     &                       $T(s)$&       0.3228&          0.4017&      0.8719&     0.0630 \\ \hline
   \multirow{3}*{Bunny}&  \multirow{3}*{54.51\%}&  \multirow{3}*{8051}&      \multirow{3}*{0.0608}&     \multirow{3}*{7205}&    ${\varepsilon _{\bf{R}}}$&       0.2013&          0.2356&      0.7826&     0.0765 \\
                       &                        &                  &                      &                     &    ${\varepsilon _{\vec t }}$&      0.1714&          0.2076&      1.0398&     0.0666 \\
                       &                        &                  &                      &                     &                        $T(s)$&      1.1260&          1.4157&      2.4135&     0.1861 \\ \hline
\multirow{3}*{Armadillo}& \multirow{3}*{61.93\%}&  \multirow{3}*{3846}&      \multirow{3}*{0.0632}&     \multirow{3}*{5558}&     ${\varepsilon _{\bf{R}}}$&      0.1514&          0.1534&      0.4626&     0.0713 \\
                       &                        &                  &                      &                     &    ${\varepsilon _{\vec t }}$&      0.0670&          0.0644&      0.7917&     0.0685 \\
                       &                        &                  &                      &                     &                        $T(s)$&      0.4484&          0.5147&      2.1055&     0.0871 \\ \hline
\multirow{3}*{Angel}&     \multirow{3}*{72.26\%}&  \multirow{3}*{12999}&      \multirow{3}*{0.5144}&     \multirow{3}*{13540}&     ${\varepsilon _{\bf{R}}}$&      0.0122&          0.0112&      0.0014&     0.0014 \\
                       &                        &                  &                      &                     &    ${\varepsilon _{\vec t }}$&      0.3873&          0.3843&      0.0382&     0.0307 \\
                       &                        &                  &                      &                     &                        $T(s)$&      1.8085&          1.7264&      2.5578&     1.5934 \\ \hline

\end{tabular} }
\end{table}

\subsection{Application}

\begin{figure*}[!t]
  \centering
  \centerline{\includegraphics[scale=0.7]{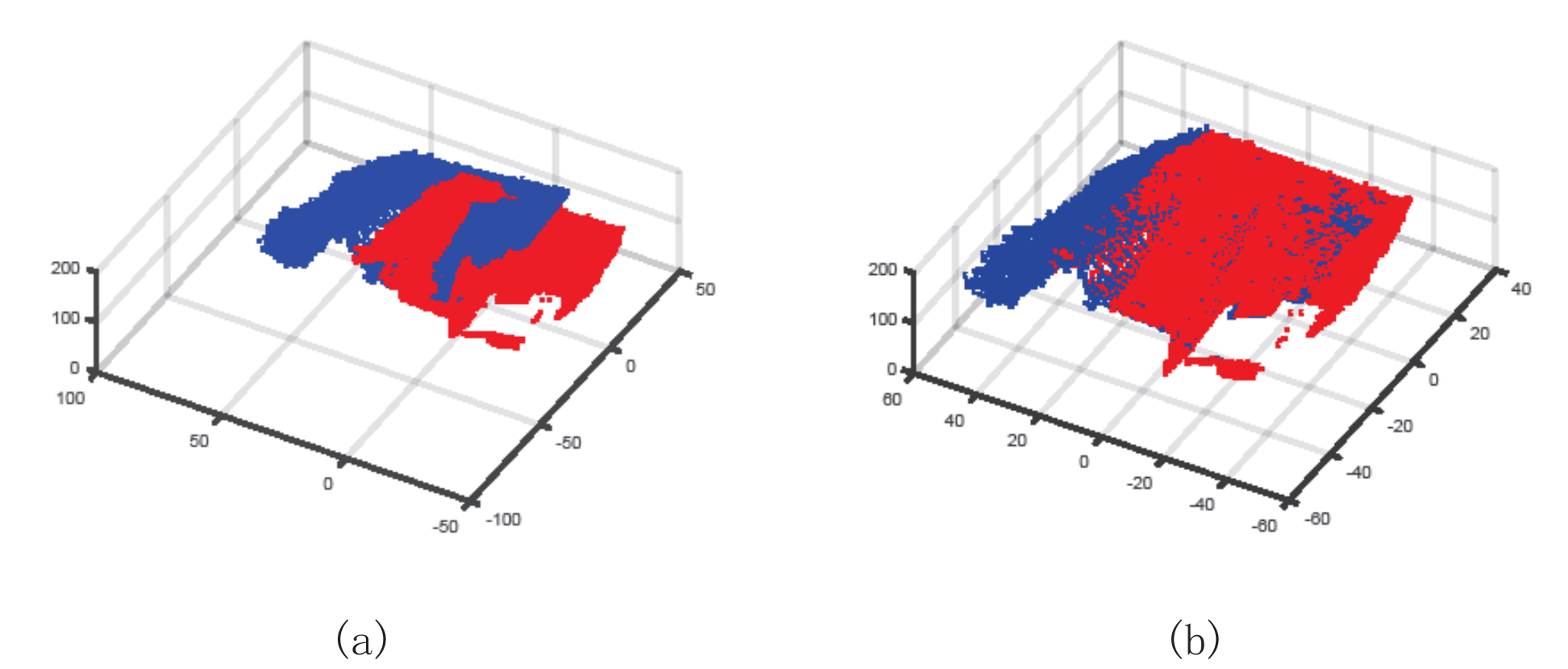}}
\caption{Application of the proposed approach to environmental point clouds. (a) Point clouds to be registered. (b) Our results. }\medskip
\label{fig:envirsets}
\end{figure*}

To confirm its application to environmental scenes, the proposed approach was testified in the following large-scale environmental point clouds, which were recorded in the Institute of Robotics and Mechatronics at the German Aerospace Center (DLR) \cite{EnvirSets}. For conducting this experiment, the given pose was imposed by a perturbed transformation $\left( {{{\bf{R}}_r},{{\vec t }_r}} \right)$, and then used as the initial transformation. Note that the given pose was obtained via some methods and are available online. Under such deployment, the registration results using the proposed approach can be demonstrated in Fig. \ref{fig:envirsets}.

As illustrated in Fig. \ref{fig:envirsets}, the proposed approach is appliable to environmental point clouds and is able to achieve pretty precise registration, even if the point cloud contains tens of millions points. This is easy to interpret because of the hard and soft assignments combined method. On one hand, the hard assignment exclude most outliers from registration. Therefore, the reconstruction of the environmental scene is mainly based on inliers, which are essential for registration. On the other hand, the soft assignment strategy assigns a value for each pair of correspondence to denote its reliability of the established correspondence. This method takes the probability into consideration and overcomes the disadvantage of merely optimizing the least square function, thus can achieve more precise registration.

\section{Conclusion}
This paper proposes a novel approach for registration of partially overlapping point clouds. The novelty of this approach is the introduction of both the hard and soft assignments, which can be represented as a binary variable and continuous variable, respectively. By consideration of the forward distance, each point in the data shape can be assigned a binary value, which can denote whether it is an inlier or outlier. Meanwhile, based on the ratio of bidirectional distances, each established point correspondence can be assigned a continuous probability, which can be viewed as an index to indicate the reliability of point correspondence. By integrating both the hard and soft assignments, a new objective function can be designed for registration of partially overlapping point clouds and a variant of the ICP algorithm is proposed to calculate the optimal rigid transformation. The proposed approach has been tested on public datasets and experimental results illustrate the proposed approach can achieve the registration of partially overlapping clouds with good robustness and accuracy.

\section*{Acknowledgements}

This work is supported by the National Natural Science Foundation of China under Grant Nos. 61573273, 61573280 and 61503300.


\bibliography{mybibfile}
\bibliographystyle{mybibfile}

\end{document}